\documentclass[runningheads]{llncs}
\usepackage[T1]{fontenc}
\usepackage{amsmath} 
\usepackage{epsfig}
\usepackage{amssymb}
\usepackage{graphicx}
\usepackage{booktabs} 
\usepackage{hyperref}
\usepackage{marvosym}
\begin{document}
\title{ShapeMamba-EM: Fine-Tuning Foundation Model with Local Shape Descriptors and Mamba Blocks for 3D EM Image Segmentation}
\titlerunning{ShapeMamba-EM}
%
\author{Ruohua Shi\inst{1,2,3} \and
Qiufan Pang\inst{1} \and
Lei Ma\inst{1,2,3} \and
Lingyu Duan\inst{1,2,4} \and
Tiejun Huang\inst{1,2,3} \and
Tingting Jiang\inst{1,2,3}\textrm{\Letter}
}

%
\authorrunning{R. Shi et al.}
%
\institute{
National Engineering Research Center of Visual Technology, School of Computer Science, Peking University, Beijing, China \and
State Key Laboratory of Multimedia Information Processing, School of Computer Science, Peking University, Beijing, China \and 
National Biomedical Imaging Center, Peking University, Beijing, China \and 
Peng Cheng Laboratory, Shenzhen, China
\\
\email{\{shiruohua@,pqf@stu.,lei.ma@,lingyu@,tjhuang@,ttjiang@\}pku.edu.cn}
}

%
%
\maketitle              
\begin{abstract}
Electron microscopy (EM) imaging offers unparalleled resolution for analyzing neural tissues, crucial for uncovering the intricacies of synaptic connections and neural processes fundamental to understanding behavioral mechanisms. Recently, the foundation models have demonstrated impressive performance across numerous natural and medical image segmentation tasks. However, applying these foundation models to EM segmentation faces significant challenges due to domain disparities. This paper presents \textit{ShapeMamba-EM}, a specialized fine-tuning method for 3D EM segmentation, which employs adapters for long-range dependency modeling and an encoder for local shape description within the original foundation model. This approach effectively addresses the unique volumetric and morphological complexities of EM data. Tested over a wide range of EM images, covering five segmentation tasks and 10 datasets, ShapeMamba-EM outperforms existing methods, establishing a new standard in EM image segmentation and enhancing the understanding of neural tissue architecture.

\end{abstract}
\section{Introduction}

Electron microscopy (EM) allows the imaging of neural tissue at a resolution sufficient to resolve individual synapses and fine neural processes. Therefore, the segmentation of EM images plays a pivotal role in the realm of biological research, offering profound insights into the inner mechanisms underlying behavior and helping drive future theoretical experiments~\cite{schneider2016quantitative,motta2019dense,hulse2021connectome}. Many excellent algorithms and datasets have emerged~\cite{isbi2012,isensee2021nnu,wei2020mitoem,shi2022u,shi2024evidential,shi2023ps}.

Recent advancements in computer vision have spurred breakthroughs by the foundation models, such as the Segment anything model (SAM), which has achieved promising zero-shot segmentation performance on a variety of natural image datasets~\cite{kirillov2023segment}. However, their performance significantly declines when applied to some downstream tasks, like biological and medical images, primarily due to the substantial disparity between the two image domains.

An effective method to enhance the generalizability of the foundation models to downstream tasks lies in fine-tuning. Recently, numerous methods applied to medical datasets have demonstrated impressive efficacy~\cite{chen2023ma,wang2023sam,zhang2023segment,zhang2024segment}. For instance, Med-SA~\cite{wu2023medical} maintains the pre-trained SAM parameters frozen while integrating LoRA modules to the designated positions. MedSAM~\cite{ma2024segment} has achieved this by fine-tuning the decoder with 1.1 million masks, enabling SAM’s application in medical imaging. In addition to these methods trained on 2D medical images, some methods propose solutions for 3D images. 3DSAM-adapter~\cite{gong20233dsam} proposes a holistically designed scheme for transferring SAM from 2D to 3D for promptable medical image segmentation. SAM-Med3D~\cite{wang2023sam} reformulates SAM to a thorough 3D architecture trained on a comprehensively processed large-scale volumetric medical dataset.

\begin{figure}[t]
\includegraphics[width=\textwidth]{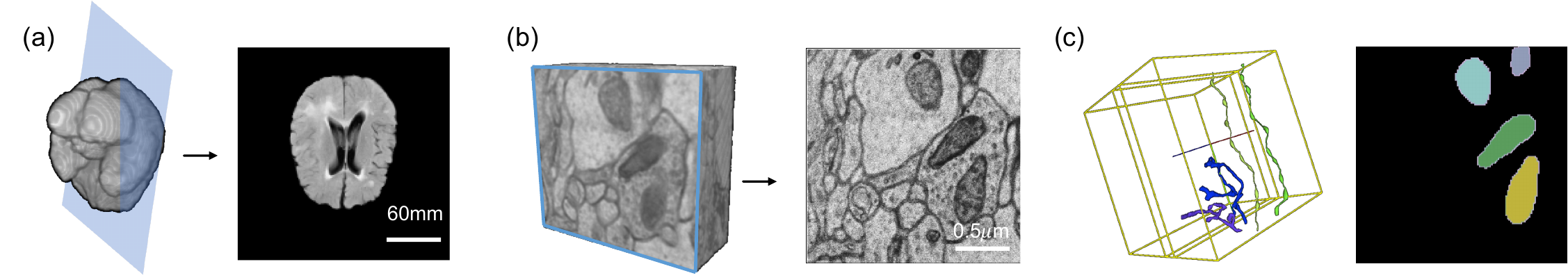}
\caption{Illustration of the medical data and EM data. (a) MRI edema image from the BraTS2021 dataset. (b) EM mitochondria data from MitoEM-R dataset. (c) 3D and 2D segmentation results of (b). The boundaries of instances share a similar local shape, and the scope of the instance spans the entire volume.} 
\label{fig:example}
\end{figure}



However, the medical foundation models above cannot be directly applied to EM data. Taking Fig.~\ref{fig:example} as an example, although both medical and EM images are all volumetric grayscale images, EM images have significantly higher resolution (about $10^5$ times), which results in more noise. Besides, the objects to be segmented in EM images have relatively consistent local features and are widely distributed across the spatial domain. This distribution, combined with the inherent anisotropy, intensifies the challenges of EM segmentation. Consequently, EM images require customized methods to address these challenges. On the other hand, medical images have grayscale and volumetric data, which are closer to EM images than to natural images, so the fine-tuned medical SAM models are more suitable for EM data than SAM. Therefore, this paper proposes a fine-tuning method specifically designed for EM image segmentation, based on a 3D medical foundation model, named \textbf{ShapeMamba-EM}. ShapeMamba-EM first modifies the original foundation model to enhance the tuning efficiency, and then adds two novel modules targeting the local morphological features and long-range dependencies in EM data as shown in Fig.~\ref{fig:example} (b,c). 


Specifically, for the selection of the foundation model, we opted for the currently largest model trained on 3D medical images, SAM-Med3D~\cite{wang2023sam}, which consists of three parts: image encoder, mask decoder, and prompt encoder. Inspired by existing work on medical data, we leverage FacT~\cite{jie2023fact} to tune the image encoder module for retaining most pre-trained weights while only updating lightweight weight increments. Furthermore, the prompt encoder is removed because crafting appropriate prompts is a challenging task for EM data and automatic segmentation has shown promise.



Besides modifying the original model, we tackle the challenges of \textit{local morphological features} and \textit{long-range dependencies} for EM data with two novel modules. Firstly, we find pixel-wise prediction alone insufficient. To accurately segment objects with similar \textit{local morphological features} and imperfect edges, we implement a 3D U-Net architecture~\cite{cciccek20163d} to predict the Local Shape Descriptors (LSDs)~\cite{sheridan2023local} to enhance the boundary prediction. Secondly, to address the challenge of \textit{long-range dependency} inherent in EM object analysis, we draw inspiration from the recent innovation in Mamba~\cite{gu2023mamba} which utilizes state space sequential models (SSMs)~\cite{gu2021efficiently,ma2024u}. These models excel at extracting long-dependency information with reduced computational burden and lower memory consumption. Consequently, we propose the integration of \textit{3D Mamba Adapters} into the image encoder. Through extensive experimentation, we demonstrate the superior performance of the ShapeMamba-EM framework across a broad spectrum of EM images, spanning five segmentation tasks and 10 datasets.

\section{Method}

\subsection{Overview}
\begin{figure}[t]
\includegraphics[width=\textwidth]{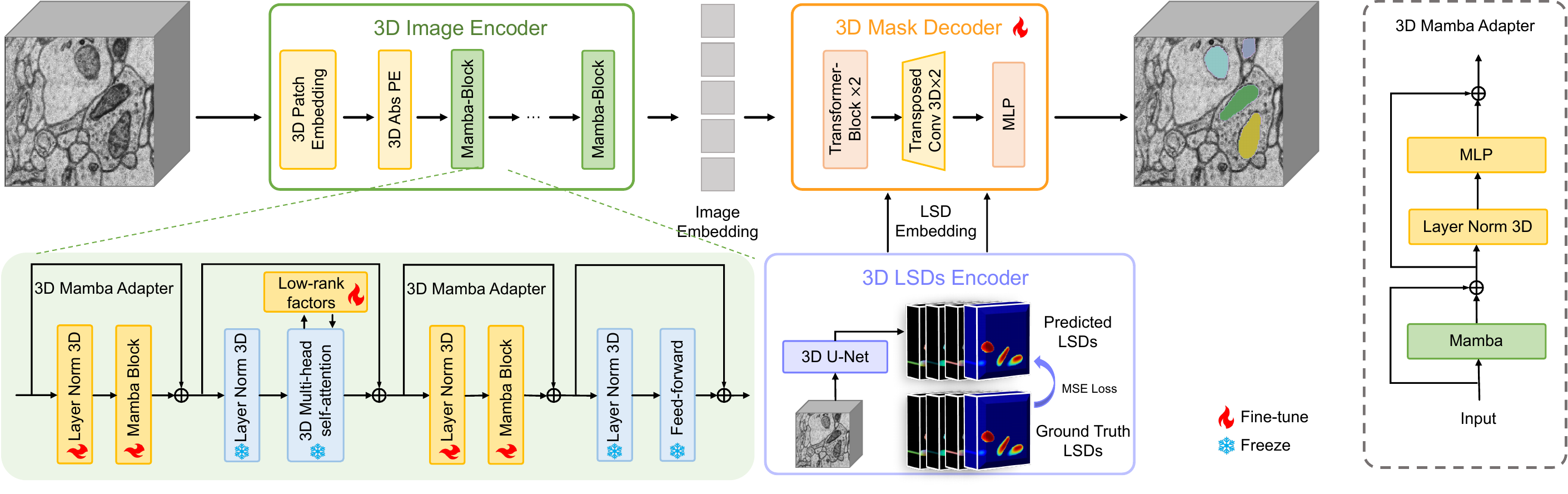}
\caption{The overall architecture of ShapeMamba-EM. The image encoder is updated with FacT. The volumetric or temporal information is effectively incorporated via a set of 3D Mamba adapters. The mask decoder is fully fine-tuned and modified to recover the prediction resolution. The LSDs are trained by the 3D U-Net network.} 
\label{fig:method}
\end{figure}


The overall framework of our proposed ShapeMamba-EM is illustrated in Fig.~\ref{fig:method}. We introduce an innovative model that enhances the SAM-Med3D core framework for 3D EM segmentation tasks. Specifically, we have augmented the SAM-Med3D architecture by incorporating the FacT approach into the 3D Multi-head self-attention model. This integration enables a more effective and efficient fine-tuning process. Furthermore, we introduce \textit{3D Mamba Adapters} designed to tackle the challenges of long-range dependency of segmentation objects. Besides, a 3D U-Net network is incorporated into the \textit{3D Mask Decoder} to capture the  Local Shape Descriptors of the segmentation objects. These two designs address the inherent limitations of SAM-Med3D in EM segmentation, thereby increasing segmentation accuracy and efficiency.

\subsection{SAM-Med3D}
In this paper, we adopt SAM-Med3D~\cite{wang2023sam} as our foundation model, due to its excellent ability to extract volumetric data features by training on a large amount of 3D medical data. It is designed based on the segment anything model (SAM)~\cite{kirillov2023segment} for volumetric medical imaging by incorporating a 3D structure for direct spatial information capture. SAM-Med3D employs a 3D convolution for patch embedding, uses a 3D positional encoding extended from SAM's 2D version, and inputs these to 3D attention blocks. These blocks, enhanced with 3D relative positional encodings, are part of the Multi-Head Self-Attention module, facilitating direct spatial detail capture. The prompt encoder uses 3D position encodings for sparse prompts and 3D convolutions for dense prompts, while the 3D Mask Decoder integrates 3D upscaling with 3D transposed convolution.

\subsection{Parameter-efficient fine-tuning of 3D image encoder}
In order to effectively extract image features, Med-SAM3D’s 3D Image Encoder comprises a substantial portion of network parameters. Fine-tuning all these weights is computationally intensive. Previous research has shown that PETL techniques can achieve adaptation performance similar to full fine-tuning but with significantly fewer network parameters updated~\cite{qiao2023vln,jie2023fact,tu2023visual}. In this regard, we adopt FacT~\cite{jie2023fact}, a SOTA parameter efficient transfer learning (PETL) technique, that can obtain comparable or superior performance compared to other PETL methods while introducing a smaller number of trainable parameters.

\subsection{3D Mamba Adapter}

In 3D EM images, the objects are widely distributed yet dense. To address this long-range dependencies challenge in EM images, we concentrate on enhancing the representational capacity of the encoder in neural network architectures through the incorporation of the Mamba layer, which is designed to capture long-range dependencies within the input data. Specifically, we design the \textit{3D Mamba Adapter} as illustrated in Fig.~\ref{fig:method} (right), which substitutes the self-attention module in the transformer architecture with the more efficient Mamba layer. Each 3D Mamba Adapter consists of a \textit{3D Layer Norm} operation with a \textit{Mamba Block}~\cite{gu2023mamba}. This enables both multi-scale and global feature modeling while maintaining a high efficiency during training and inference. 

\subsection{3D Local Shape Descriptor Encoder}

\begin{figure}[t]
\includegraphics[width=\textwidth]{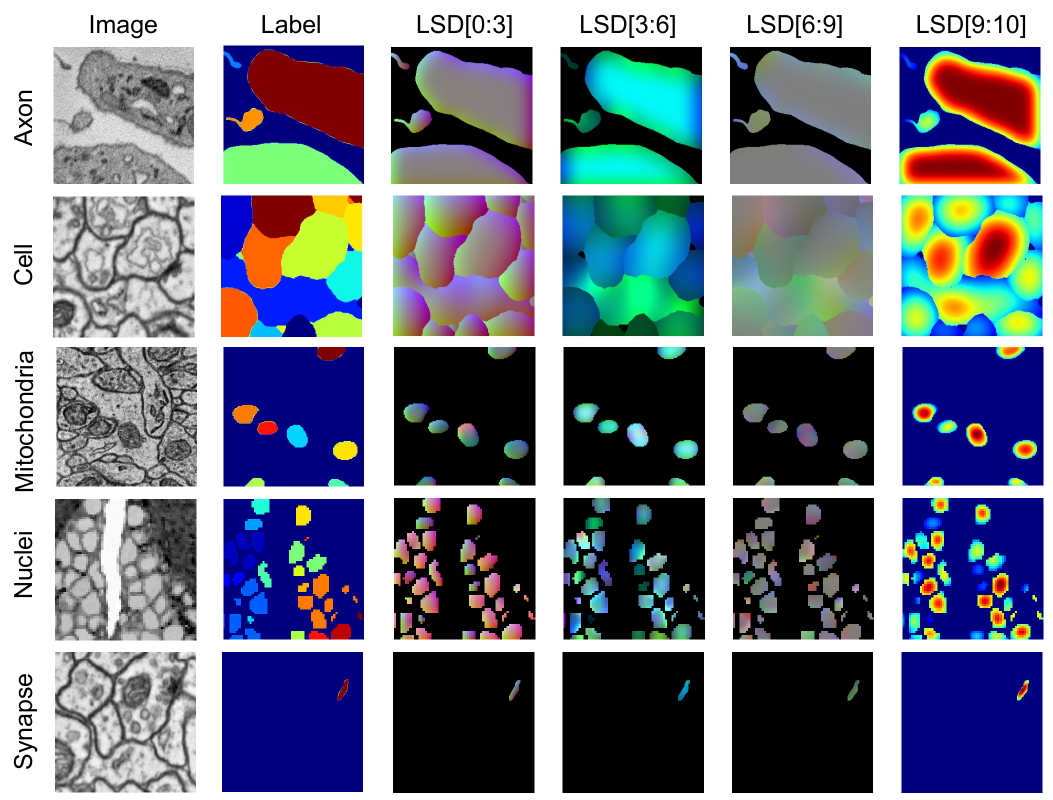}
\caption{Visualizations of LSDs for different segmentation tasks. From left to right:  the EM image, segmentation labels, and components of the LSDs.} \label{fig:lsd}
\end{figure}

Although recent deep learning networks have made good progress in EM image segmentation, a neural network might focus only on a few center voxels to detect objects and achieve high accuracy during training, especially if trained using a voxel-wise loss. However, this strategy might fail in rare cases where boundary evidence is ambiguous. The objects to be segmented in EM images, such as axons, cells, and mitochondria, have regular local shapes. Inspired by~\cite{sheridan2023local}, to enhance the model's learning of the local shapes objects in EM images, We first define the Local Shape Descriptors (LSDs) of a segment $i \in\{1, \ldots, l\}$ under a given voxel $v$. we intersect the segment $\mathrm{y}(v)$ underlying a voxel $v \in \Omega$ with a 3D ball of radius $\sigma$ centered at $v$ to obtain a subset of voxels $S_v \subset \Omega$, formally given as:
\begin{equation}
S_v=\left\{v^{\prime} \in \Omega\left|\mathrm{y}(v)=\mathrm{y}\left(\nu^{\prime}\right),\right| v-\left.v^{\prime}\right|_2 ^2 \leq \sigma\right\} .
\end{equation}
The LSD $\operatorname{lsd}^{\mathrm{y}}: \Omega \mapsto \mathbb{R}^{10}$ for a voxel $v$ is a concatenation of the size, center offset and coordinate covariance, that is:
\begin{equation}
\operatorname{lsd}^{\mathrm{y}}(\boldsymbol{v})=(\underbrace{\mathrm{s}\left(S_v\right),}_{\text {size }} \underbrace{\mathrm{m}\left(S_v\right)-v}_{\text {center offset }}, \underbrace{\mathrm{c}\left(S_v\right)}_{\text {covariance }}) .
\end{equation}
where $\mathrm{s}\left(S_v\right)=\left|S_v\right|$ is the size of $S_v$, $\mathrm{m}\left(S_v\right)$ is the covariance of its coordinates and $\mathrm{c}\left(S_v\right)$ is the mean coordinates:
\begin{equation}
\mathrm{m}\left(S_v\right)=\frac{1}{\mathrm{~s}\left(S_v\right)} \sum_{v \in S_v} v ,\quad
\mathrm{c}\left(S_v\right)=\frac{1}{\mathrm{~s}\left(S_v\right)} \sum_{v \in S_v}\left(v-\mathrm{m}\left(S_v\right)\right)\left(v-\mathrm{m}\left(S_v\right)\right)^{\mathrm{T}}.
\end{equation}

Therefore, for a 3D image, LSDs are represented as a ten-dimensional embedding. This encapsulation includes: LSD[0:3] for the mean offset; LSD[3:6] and LSD[6:9] delineating the covariance of coordinates, with LSD[3:6] capturing the diagonal entries and LSD[6:9] the off-diagonals; and finally, LSD[9:10] reflecting the size, quantified as the number of voxels within the intersected Gaussian. Some visualization examples of LSDs are shown in Fig.~\ref{fig:lsd}. We use lsd$^y(v)$ to formulate an auxiliary learning task that complements the prediction of affinities. For that, we use the 3D U-Net network to learn the lsd$^{\mathrm{x}}: \Omega \mapsto \mathbb{R}^{10}$ directly from raw data $\mathrm{x}$, and take it as the embedding to the 3D Emage Decoder network. 




\section{Experiments and Results}
We extensively evaluate our method on five EM image segmentation tasks, covering 10 datasets, i.e., axon segmentation, cell segmentation, mitochondria segmentation, synapse segmentation, and nuclei segmentation. We first conduct comparisons with state-of-the-art EM image segmentation methods and SAM fine-tuning methods, and then provide ablation studies to analyze our method.

\begin{table}[t]
    \centering
    \begin{tabular}{c|c|c|c|c|c|c|c|c|c|c|c|c}
    \toprule \textbf{Task} &\multicolumn{2}{|c|}{\textbf{Axon}}  &\multicolumn{6}{|c}{\textbf{Cell}} & \multicolumn{4}{|c}{\textbf{Synapse}} \\
    \toprule Dataset & \multicolumn{2}{|c|}{Gauy} & \multicolumn{2}{|c|}{ISBI2012} & \multicolumn{2}{|c|}{SNEMI3D} & \multicolumn{2}{|c}{CREMI}& \multicolumn{2}{|c|}{CREMI} & \multicolumn{2}{|c}{EM-R50} \\
    \toprule Metrics & Dice & mAP & Dice & mAP & Dice & mAP & Dice & mAP & Dice & mAP & Dice & mAP  \\
    \toprule U3D-BCD &0.790 &0.801 &0.932 &0.930  &0.964 &0.961 & 0.952 &0.960 &0.814 &0.823 &0.764 &0.775\\
    \hline SwinU &0.722 &0.746  &0.932 & 0.935 &0.962 &0.958 &0.943&0.950 & 0.821& \underline{0.829}&0.733 &0.749\\
    \hline nnU-Net &0.788 &0.789  &\underline{0.971} & \underline{0.966} &\underline{0.965} &\underline{0.962} &0.947&0.951 &\underline{0.828} &0.814&0.760 &0.758 \\
    \hline MA-SAM &0.539 &0.621  &0.763 & 0.742 &0.874 &0.878 &0.836&0.845 &0.475 & 0.464 &0.527 &0.539 \\
    \hline 3DSAMA & 0.474&0.599  & 0.518 &0.569  &0.371 &0.428 &0.531 &0.594 &0.01 &0.01 &0.01 &0.01 \\
    \hline MSAM3D &0.716 &0.743  &0.858 &0.862  &0.931 &0.927 &0.943 &0.948 &0.611 &0.591 &0.711 & 0.718\\
    \hline w/o M &0.735 &0.750  &0.947 &0.951  &0.962 &0.965 &\underline{0.954} &\underline{0.966} & 0.815& 0.823 &0.779 &0.780 \\
    \hline w/o L &\underline{0.791} &\underline{0.796}  &0.939 &0.942  &0.959 &0.958 &0.942 &0.951 &0.799& 0.810&\underline{0.780} &\underline{0.801} \\
    \hline Ours & \textbf{0.809} & \textbf{0.827} & \textbf{0.958}& \textbf{0.951}  & \textbf{0.974} & \textbf{0.972}&\textbf{0.965}& \textbf{0.973} &\textbf{0.834} &\textbf{0.865} &\textbf{0.792} &\textbf{0.817}\\
    \toprule
    \end{tabular}
    \caption{Quantitative outcomes of methods applied to segmentation tasks for axons, synapses, and cells. Bold and underlined numbers denote the 1st and 2nd scores.}
    \label{tab:result1}
\end{table}
\begin{table}[t]
    \centering
    \begin{tabular}{c|c|c|c|c|c|c|c|c|c|c|c|c}
    \toprule \textbf{Task} & \multicolumn{10}{|c}{\textbf{Mitochondria}} & \multicolumn{2}{|c}{\textbf{Nuclei}}\\
    \toprule Dataset & \multicolumn{2}{|c|}{Gauy}& \multicolumn{2}{|c|}{Kasthuri++}& \multicolumn{2}{|c|}{Lucchi++}& \multicolumn{2}{|c|}{MitoEM-H}& \multicolumn{2}{|c|}{MitoEM-R} & \multicolumn{2}{|c}{NucMM-Z}\\
    \toprule Metrics  & Dice & mAP& Dice & mAP& Dice & mAP & Dice & mAP & Dice & mAP & Dice & mAP \\
    \toprule U3D-BCD &0.564 &0.529 &0.889 & 0.831 & 0.880& 0.753& 0.746 & 0.773& 0.775 & 0.844&0.879 &0.894\\
    \hline SwinU &0.472 &0.443 & 0.904 & 0.861& 0.869& 0.874& 0.779& 0.822&0.803 & \underline{0.867}& 0.866 &0.837\\
    \hline nnU-Net &0.528 &0.501 & 0.859 & 0.872& 0.856& 0.829& 0.807 & 0.830& 0.825 & 0.864&0.907 &0.894\\
    \hline MA-SAM &0.349 &0.315 &0.769 &0.773 & 0.754&0.740 & 0.692& 0.701&0.723 & 0.744& 0.839&0.858\\
    \hline 3DSAMA &0.145 &0.188 &0.582 &0.668 &0.645 &0.682 &0.671 &0.710 &0.786 &0.802 &0.863 &0.897\\
    \hline MSAM3D &0.537 &0.521 &0.902 &0.878 &0.715 &0.728 &0.711 & 0.714&0.822 &0.835 &0.878 &0.895\\
    \hline w/o M &0.572 &\underline{0.598} &0.951 &0.920 &0.906 & \underline{0.910}&0.817 &\underline{0.825} &\underline{0.846} &0.852 &\underline{0.910} &0.903\\
    \hline w/o L &\underline{0.586} &0.591 &0.942 &\underline{0.922} &\underline{0.915} &0.908 & \underline{0.820}&0.812 &0.839 &0.844 &0.904 &\underline{0.899}\\
    \hline Ours &\textbf{0.612 } &\textbf{0.603}  &\textbf{0.968}&\textbf{0.936} &\textbf{0.940} & \textbf{0.954} &\textbf{0.847} &\textbf{0.877} &\textbf{0.852} &\textbf{0.930} &\textbf{0.915} &\textbf{0.907}\\
    \toprule
    \end{tabular}
    \caption{Quantitative outcomes of methods applied to segmentation tasks for mitochondrion and nucleus. Bold and underlined numbers denote the 1st and 2nd scores.}
    \label{tab:result2}
\end{table}

\subsection{Datasets and Experimental Settings}
We conduct a series of extensive experiments on 5 segmentation tasks with 10 datasets to evaluate the performance of our method. Here we briefly introduce them, and more details are shown in the supplementary materials. Specifically, the \textit{Axon Segmentation Task} uses the Gauy dataset~\cite{guay2021dense}, the \textit{Cell Segmentation Task} uses three datasets: ISBI2012~\cite{isbi2012}, SNEMI3D~\cite{lee2017superhuman} CREMI~\cite{cremi}, the \textit{Mitochondira Segmentation Task} uses five datasets: MitoEM-R~\cite{wei2020mitoem}, MitoEM-H~\cite{wei2020mitoem}, Kasthuri++~\cite{casser2020fast}, Lucchi++~\cite{casser2020fast}, and Gauy~\cite{guay2021dense}, the \textit{Nuclei Segmentation Task} uses NucMM-Z~\cite{lin2021nucmm} dataset, ans the \textit{Synapse Segmentation Task} uses two datasets: CREMI~\cite{cremi} and EM-R50~\cite{lin2020two}. For the evaluation, we evaluate the methods using mean 3D Average Precision (mAP)~\cite{wei2020mitoem} and Dice scores at the instance level.

We comprehensively compare our proposed method against a suite of cutting-edge algorithms. These include recent successful approaches for biomedical data segmentation utilizing CNN architectures, such as U3D-BCD~\cite{wei2020mitoem} and nnU-Net~\cite{isensee2021nnu}, alongside Transformer-based architecture, SwinUNETR~\cite{hatamizadeh2022swin} (SwinU). Additionally, we examine fine-tuning methods based on SAM, specifically MA-SAM~\cite{chen2023ma} and 3DSAM-Adapter~\cite{gong20233dsam} (3DSAMA). Our evaluation also extends to direct fine-tuning using Med-SAM3D (MSAM3D), supplemented by an ablation study to assess the impact of our novel 3D Mamba Adapter and 3D LSD Encoder Module. During the training of fine-tuning methods, we independently trained the model for each task and employed Binary Cross-Entropy (BCE) loss during training. The experiments are conducted training on 8 NVIDIA A800 GPUs. More details of the experiments including the splits of the training and testing data, $\sigma$ in LSD generation are shown in the supplementary materials.

\subsection{Quantitative and qualitative segmentation results}

\textbf{Qualitative results}. The quantitative results in Tab.~\ref{tab:result1} and Tab.~\ref{tab:result2} underscore the superior performance of our proposed method in precise EM segmentation across all five tasks. Compared to both CNN-based, Transformer-based, and fine-tuning based methods, the proposed ShapeMamba-EM demonstrates competitive even higher performance. We showcase several predictions in Fig.~\ref{fig:result}. More comparisons of other SOTA methods for each dataset are provided in the supplementary material.

\textbf{Comparisions of fine-tuning methods.} In the comparative analysis of fine-tuning algorithms, it indicates that those based on Med-SAM3D reflects significant improvement over those based solely on SAM (such as MA-SAM and 3DSAM-Adapter). Such progress underscores the effectiveness of leveraging medical data in refining the model's ability to generalize to EM data. In Fig.~\ref{fig:result}, we observe that SAM-based methods demonstrate poor performance in segmenting small objects and dense cells, which strongly requires volumetric spatial information. Particularly for the 3D SAM-Adapter method, it seldom predicts the synapse. This limitation might stem from its reliance on the quality of prompts during training. Considering that the number of positive prompts in synapse datasets is substantially lower than that of negative prompts, the model faces challenges in learning useful information effectively. Furthermore, its performance in cell segmentation is hindered by the static number of prompts.

\textbf{Ablation study for 3D Mamba Adapter and 3D LSD Encoder.} We further evaluate the effectiveness of the 3D Mamba Adapter (w/o M) and the 3D LSD Encoder (w/o L). Results indicate an average increase of 5\% with the use of the 3D Mamba Adapter and 8\% with the 3D LSD Encoder. Additionally, we observe that the impact of the 3D Mamba Adapter on the nuclei segmentation task is relatively minor, potentially attributable to the smaller size of the images and objects to be segmented. Moreover, compared to methods that utilize multi-head self-attention modules, such as MSAM3D and 3DSAMA, the experiments confirm Mamba's superior performance. The segmentation results of the models with and without 3D Mamba Adapter are shown in the supplementary material.

\begin{figure}[t]
\includegraphics[width=\textwidth]{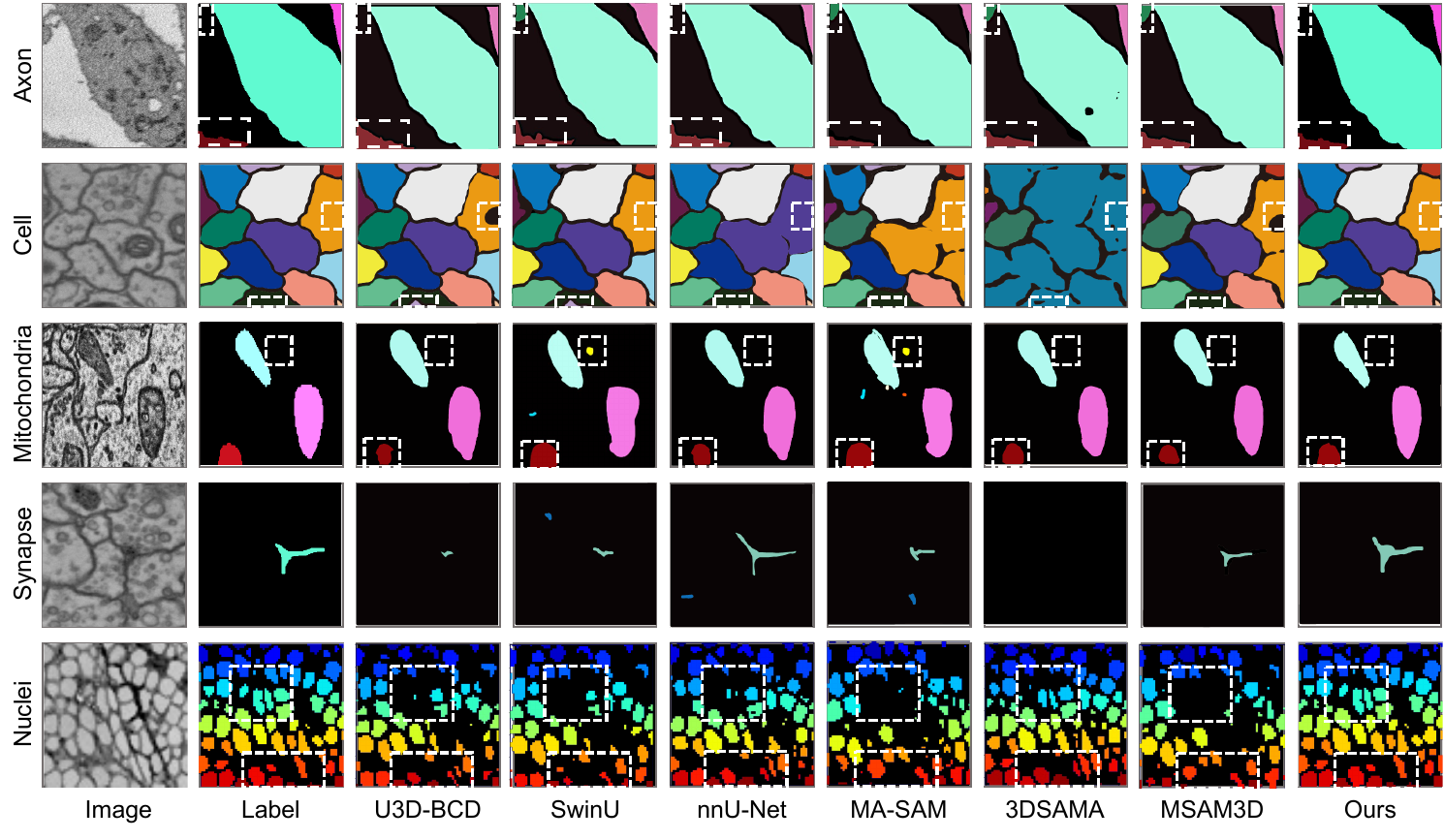}
\caption{The visualization of segmentation results.} 
\label{fig:result}
\end{figure}

%
%

\textbf{Limitations.} This paper focuses on fine-tuning foundation models for EM images. Due to space constraints, we did not compare with other Mamba-based methods as they are 2D models and do not support 3D segmentation. Additionally, we used the 3D U-Net for LSD estimation because the original method uses it. We plan to explore more advanced models in future work.

\section{Discusion and Conclusion}
In summary, ShapeMamba-EM offers an innovative fine-tuning method for EM segmentation, leveraging a 3D medical foundation model to address unique challenges of EM data such as high resolution and complex tissues. It surpasses traditional models by integrating 3D Mamba Adapters and Local Shape Descriptors Encoder, improving accuracy and efficiency. Extensive tests on diverse EM datasets highlight its effectiveness in high-resolution image segmentation, setting new benchmarks. This work aims to enhance EM segmentation and showcase the adaptability of medical foundation models for in-depth biological studies.

\begin{credits}
\subsubsection{\ackname} This work was supported by the Natural Science Foundation of China under contract 62088102, the Beijing Natural Science Foundation (Grant No. JQ24023), and the Beijing Municipal Science $\&$ Technology Commission Project (No.Z231100006623010). We also acknowledge the Biomedical Computing Platform of National Biomedical Imaging Center and High-Performance Computing Platform of Peking University for providing computational resources.

\subsubsection{\discintname} The authors have no competing interests to declare that are relevant to the content of this article.
\end{credits}

\bibliographystyle{splncs04}
\bibliography{Paper-0151}

\end{document}


\title{Supplementary for\\ShapeMamba-EM: Fine-Tuning Foundation Model with Local Shape Descriptors and Mamba Blocks for 3D EM Image Segmentation}

%
%

\author{Ruohua Shi\inst{1,2,3} \and
Qiufan Pang\inst{1} \and
Lei Ma\inst{1,2,3} \and
Lingyu Duan\inst{1,4} \and
Tiejun Huang\inst{1,2,3} \and
Tingting Jiang\inst{1,2,3}\textrm{\Letter}
}

\institute{
National Engineering Research Center of Visual Technology, School of Computer Science, Peking University, Beijing, China \and
State Key Laboratory of Multimedia Information Processing, School of Computer Science, Peking University, Beijing, China \and 
National Biomedical Imaging Center, Peking University, Beijing, China \and 
Peng Cheng Laboratory, Shenzhen, China
\\
\email{\{shiruohua@,pqf@stu.,lei.ma@,lingyu@,tjhuang@,ttjiang@\}pku.edu.cn}
}
\maketitle              

\begin{table}[h!]
    \centering
    \begin{tabular}{l|c|c|c|c|c} 
    \toprule
    \textbf{Dataset} & \textbf{Species} & \textbf{Tissue} & \textbf{Annotations} & \textbf{Image Modality} & \textbf{Resolution}\\
    \hline 
    ISBI 2012 & Drosophila & Brain-VNC  & Cell & SEM &  $50\times4\times4 nm$\\
    \hline 
    SNEMI3D & Mouse & Brain-cortex & Cell& SEM & $30\times4\times4 nm$ \\
    \hline 
    MitoEM-H & Human & Cortex & Mitochondria & mbSEM &   $30\times8\times8 nm$ \\
    \hline 
    MitoEM-R & Rat & Cortex & Mitochondria & mbSEM	&  $30\times8\times8 nm$ \\
    \hline 
    Lucchi++ & Mouse & Hippocampus & Mitochondria & FIB-SEM & 	 $5\times5\times5 nm$ \\
    \hline 
    Kasthuri++ & Mouse & Cortex & Mitochondria &ssEM & $30\times3\times3 nm$\\
    \hline 
    CREMI & Drosophila & Brain & Cell, Synapse& TEM &  $40\times4\times4 nm$ \\
    \hline 
    EMR-50 & Rat & Cortex & Synapse & mbSEM & $30\times8\times8 nm$\\
    \hline 
    Gauy & human & Platelet & Axon, Mitochondria & SBF-SEM & $50\times10\times10 nm$ \\
    \hline 
    NucMM-Z & Zebrafish & Brain & Nuclei & SEM & $30\times4\times4 nm$\\
    \hline 
    \toprule
    \end{tabular}
    \caption{Details of the datasets.}
    \label{tab:dataset_details}
\end{table}

\begin{table}[h!]
    \centering
    \begin{tabular}{l|c|c|c|c} 
    \toprule
    \textbf{Dataset} & 	\textbf{Training Set}	& \textbf{Testing Set}	& \textbf{Patch Size} & \textbf{$\sigma$}\\
    \hline 
    ISBI 2012 & 24$\times$512$\times$512	& 6$\times$512$\times$512	&100$\times$256$\times$256 & 50\\
    \hline 
    SNEMI3D & 80$\times$1024$\times$1024	& 20$\times$1024$\times$1024	& 100$\times$256$\times$256 & 50 \\
    \hline 
    MitoEM-H & $400\times4096\times4096$	& $100\times4096\times4096$	& 100$\times$256$\times$256  & 5\\
    \hline 
    MitoEM-R & $400\times4096\times4096$	& $100\times4096\times4096$	& 100$\times$256$\times$256 & 5\\
    \hline 
    Lucchi++ & 	$165\times1024\times768$ & $165\times1024\times768$ & 100$\times$256$\times$256 & 5\\
    \hline 
    Kasthuri++ & $85\times1643\times1613$	& $75\times1334\times1553$	& 100$\times$256$\times$256 & 5\\
    \hline 
    CREMI & 3$\times$125$\times$1000$\times$1000& 3$\times$125$\times$1000$\times$1000 & 100$\times$256$\times$256  & 50(cell)/5(syn)\\
    \hline 
    EMR-50 & 100$\times$700$\times$700	& 99$\times$1000$\times$1000	& 100$\times$256$\times$256 & 5\\
    \hline 
    Gauy & 74$\times$800$\times$800	& 121$\times$609$\times$400	& 100$\times$256$\times$256  & 15(axon)/5(mito)\\
    \hline 
    NucMM-Z & 18$\times$64$\times$64$\times$64	& 9$\times$64$\times$64$\times$64	& 64$\times$64$\times$64 & 2\\
    \hline 
    \toprule
    \end{tabular}
    \caption{Parameters of the experiments.}
    \label{tab:training_details}
\end{table}

\begin{table}[h!]
    \centering
    \begin{tabular}{c|c|c|c|c|c}
    \toprule
    \multirow{2}{*}{ \textbf{Dataset} } & \multirow{2}{*}{\textbf{Method}} & \multicolumn{2}{|c|}{ \textbf{Dice} } & \multicolumn{2}{|c}{ \textbf{mAP} } \\
    \cline { 3 - 6 } & & SOTA & Ours & SOTA & Ours \\
    \hline ISBI2012 & PS-Net~\cite{shi2023ps} & 0.94 & 0.958 & - & 0.951 \\
    \hline SNEMI3D & DMT~\cite{hu2021topology} & 0.971 & 0.964 & - & 0.961 \\
    \hline CREMI(Synapse) & CleftNet~\cite{liu2021cleftnet} & 0.831 & 0.834 & - & 0.865 \\
    \hline Kasthuri++ & HIVE-Net~\cite{yuan2021hive} & 0.962 & 0.968 & - & 0.936 \\
    \hline Lucchi++ & DualRel~\cite{mai2023dualrel} & 0.934 & 0.94 & - & 0.954 \\
    \hline MitoEM-H & ATFormer~\cite{chen2023atformer} & - & 0.847 & 0.782 & 0.877 \\
    \hline MitoEM-R & ATFormer~\cite{chen2023atformer} & - & 0.852 & 0.682 & 0.93 \\
    \hline NucMM-Z & U3D-BCD~\cite{lin2021nucmm} & 0.879 & 0.915 & 0.894 & 0.907 \\
    \toprule
    \end{tabular}
    \caption{Comparison with SOTA results. For each of the eight datasets, we choose the SOTA method with the best Dice or mAP performance for comparison. The performance in the original paper of the methods is reported. Since the EM-R50 and Gauy datasets lack corresponding SOTA methods reporting performance with these metrics, no SOTA methods are chosen for these two datasets.}
    \label{tab:compare_sota}
\end{table}

\begin{figure}[h!]
\centering
\includegraphics[width=0.7\textwidth]{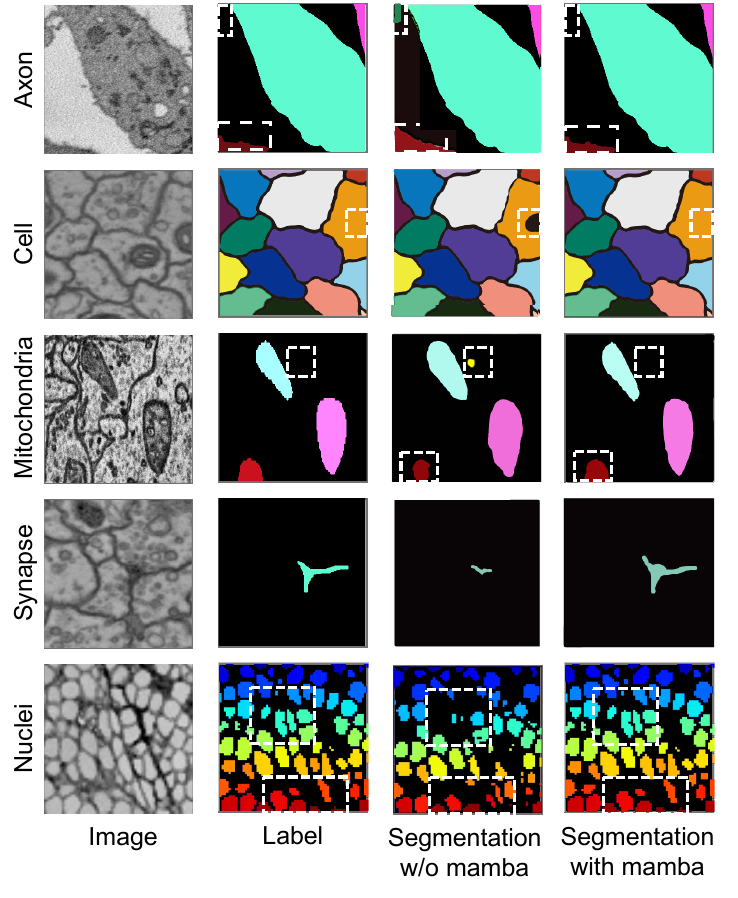}
\caption{The segmentation results of the models with and without the 3D Mamba Adapter.} 
\label{fig:result}
\end{figure}

\newpage
%
%
\bibliographystyle{splncs04}
\bibliography{Paper-0151}